\pdfoutput=1
\documentclass[11pt]{article}

\usepackage[preprint]{acl}

  \usepackage{times}
  \usepackage{latexsym}

  \usepackage[T1]{fontenc}
  \usepackage[utf8]{inputenc}
  \usepackage{CJKutf8}

  \usepackage{microtype}
  \usepackage{inconsolata}

  \usepackage{graphicx}
  \usepackage{subcaption}
  \usepackage{booktabs}
  \usepackage{array}
  \usepackage{tikz}
  \usepackage{multirow}
  \usepackage{amsmath,amssymb}
  \usepackage{url}
  \usepackage{pifont}
  \usepackage{enumitem}
  \usepackage{stfloats}
  \usepackage{float}
  \usepackage{listings}
  \usepackage{algorithm}
  \usepackage{algpseudocode}
  \usepackage{cleveref}


  \providecommand{\Cref}{\cref}

  \newcommand{\system}{\textsc{HuoziIME}}
  \newcolumntype{C}[1]{>{\centering\arraybackslash}m{#1}}
  \newcommand{\cmark}{\textcolor{green!55!black}{\ding{51}}}
  \newcommand{\xmark}{\textcolor{red!70!black}{\ding{55}}}
  \newcommand{\warnmark}{%
    \tikz[baseline=-0.6ex,x=1ex,y=1ex]{%
      \path[fill=orange!85!black,draw=orange!85!black,line join=round]
        (0,0) -- (1,1.72) -- (2,0) -- cycle;
      \node[white,font=\bfseries\tiny] at (1,0.63) {!};
    }%
  }

\title{%
\includegraphics[height=1.2em]{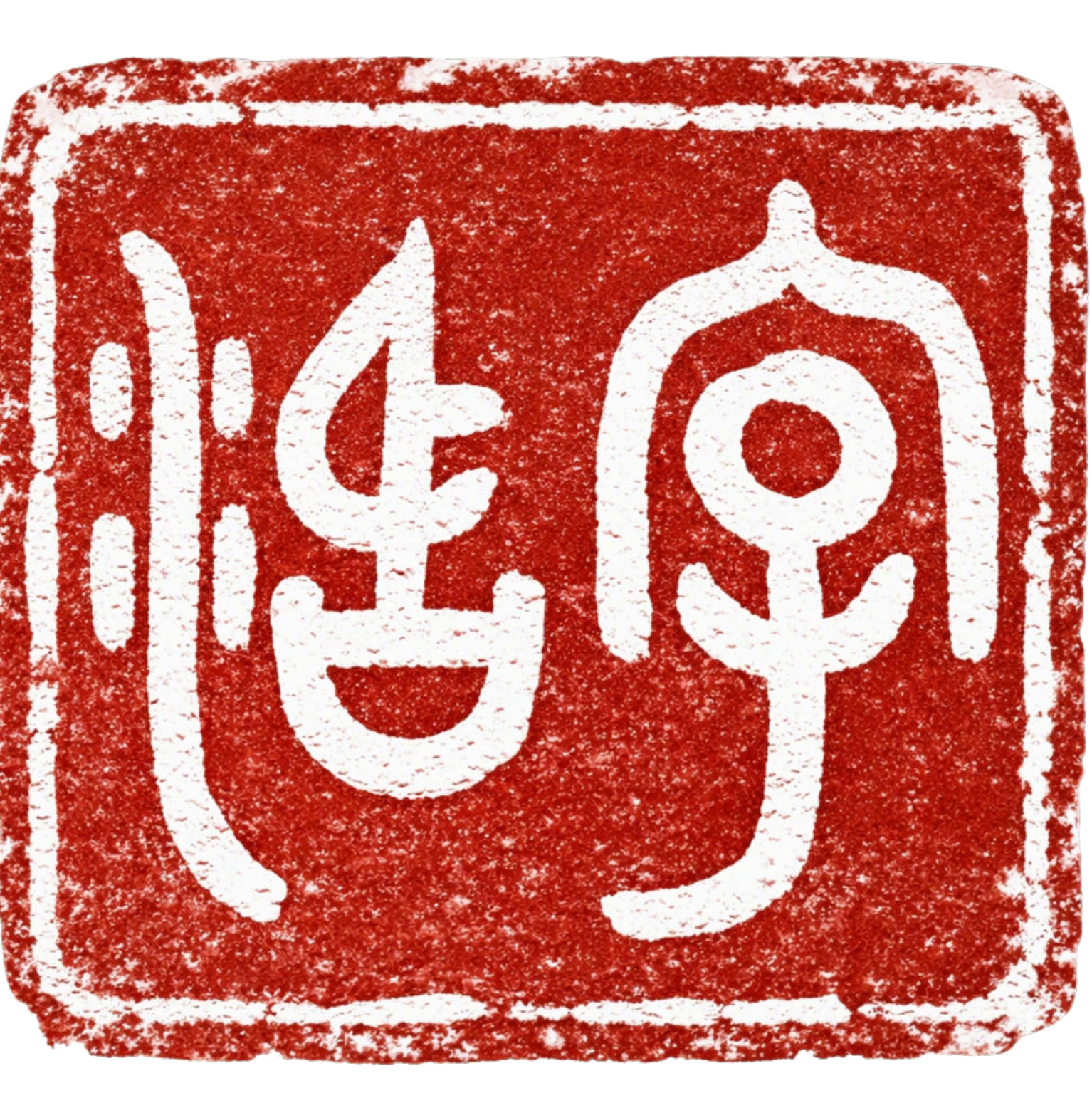}\hspace{0.35em}%
\system{}: An On-Device LLM-enhanced Input Method\\for Deep Personalization}
\author{
Baocai Shan \quad Yuzhuang Xu \quad Wanxiang Che\thanks{Corresponding author}\\
Harbin Institute of Technology, Harbin, China\\
\texttt{\{bcshan,car\}@ir.hit.edu.cn}\\
}
\date{}

\begin{document}
\maketitle

\begin{figure*}[!b]
  \centering
  \includegraphics[width=0.89\textwidth]{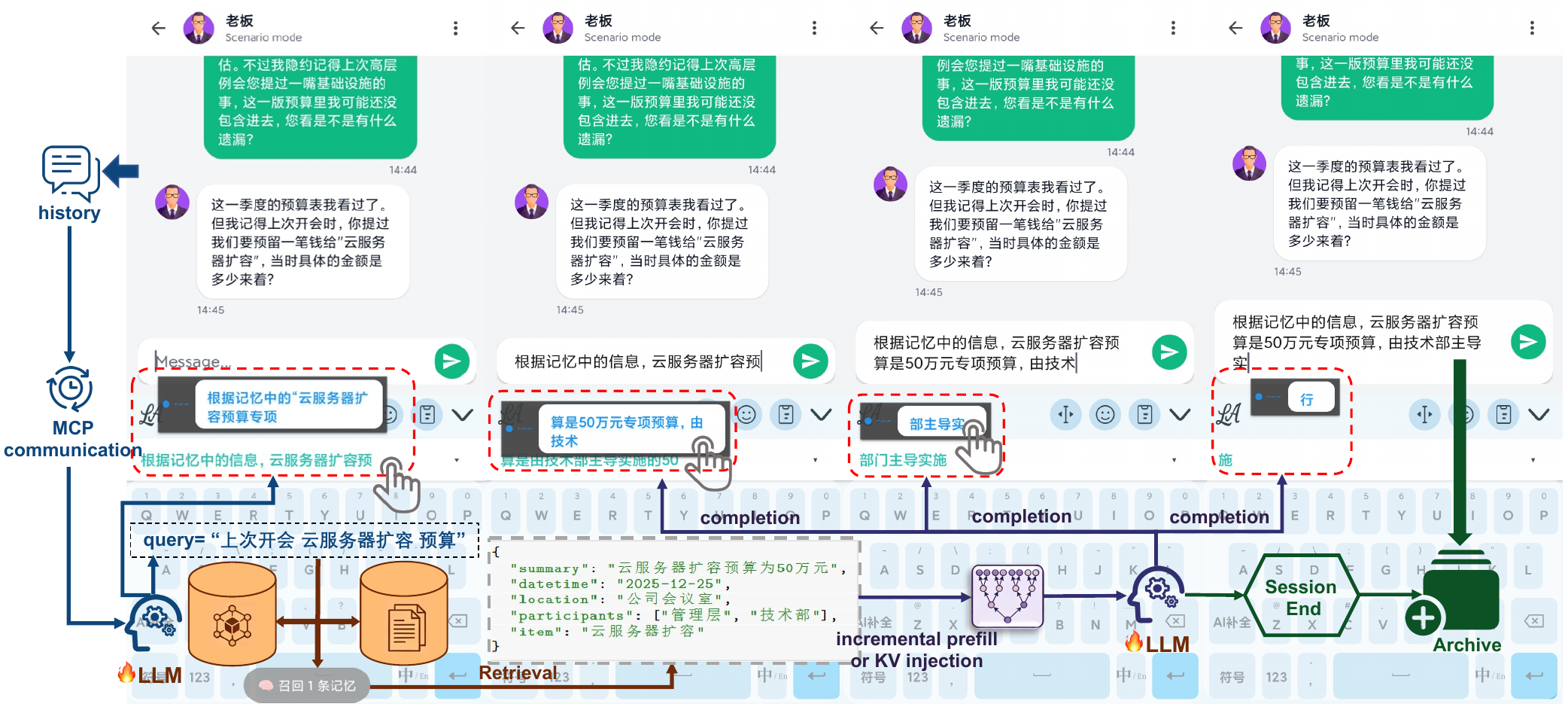}
  \caption{Overview of HuoziIME. We illustrate the end-to-end workflow, from user input and memory retrieval to LLM-based prediction, highlighting the full-process context management of LLM in IME.}
  \label{fig:teaser}
\end{figure*}

\begin{abstract}
Mobile input method editors (IMEs) are the primary interface for text input, yet they remain constrained to manual typing and struggle to produce personalized text. 
While lightweight large language models (LLMs) make on-device auxiliary generation feasible, enabling deeply personalized, privacy-preserving, and real-time generative IMEs poses fundamental challenges.
To this end, we present \system{}, a personalized on-device IME powered by LLM.
We endow \system{} with initial human-like prediction ability by post-training a base LLM on synthesized personalization data. 
Notably, a hierarchical memory mechanism is designed to continually capture and leverage user-specific input history. 
Furthermore, we perform systemic optimizations tailored to on-device LLM-based IME deployment, ensuring efficient and responsive operation under mobile constraints.
Experiments demonstrate efficient on-device execution and high-fidelity memory-driven personalization.
Code and package are available at \url{https://github.com/Shan-HIT/HuoziIME}.
\end{abstract}

  \section{Introduction}
  
  Input method editors (IMEs) serve as the most fundamental and highest-frequency interface for text entry in mobile human-device interaction. While traditional Chinese IMEs have evolved from N-gram statistical models to neural transliteration tools \cite{chen2015neural, huang2018moon}, they remain fundamentally constrained to pinyin-to-character conversion. Consequently, they struggle to model user intent or capture long-term personalized writing patterns.

  Recently, large language models (LLMs) demonstrate immense potential in writing assistance \cite{bubeck2023sparks, zhao2023survey}. Mainstream vendors begin exploring hybrid IME architectures that combine traditional candidate generation with LLM-driven intelligent writing and preset personas. However, such integrations are largely incremental: generative AI is introduced as a cloud-based auxiliary add-on rather than a foundational redesign of the IME pipeline, as shown in Table~\ref{tab:comparison}. As a result, existing IMEs remain limited in three critical aspects.

    \begin{table*}[!t]
    \centering
    \footnotesize
    \setlength{\tabcolsep}{2.6pt}
    \renewcommand{\arraystretch}{1.12}
    \resizebox{\textwidth}{!}{%
    \begin{tabular}{@{}C{0.10\textwidth}C{0.10\textwidth}C{0.10\textwidth}C{0.18\textwidth}C{0.14\textwidth}C{0.13\textwidth}C{0.19\textwidth}@{}}
      \toprule
      \textbf{Vendor} & \textbf{Model} & \textbf{Deployment} & \textbf{Core AI Functions} & \textbf{Personalization} & \textbf{Memory} & \textbf{Privacy \& Security} \\
      \midrule
      SwiftKey & GPT-4 Turbo & Cloud-first & Writing assist; context modeling; emoji gen & \xmark~Weak (static) & \xmark~No (stateless) & \warnmark~Cloud-dependent; data exposure risk \\
      BaiduIME & ERNIE Bot & Cloud-first & Writing assist; emotional chat; rewrite & \xmark~Weak (prompt-based) & \xmark~No (short-term) & \warnmark~Cloud-dependent; offline unstable \\
      SogouIME & Hunyuan & Cloud-first & Speech polish; spoken-to-written; translation & \xmark~Weak (lexicon-based) & \xmark~No (no persistent) & \warnmark~Cloud-dependent; privacy tradeoff \\
      iFlytekIME & Spark & Hybrid & Persona presets; completion; agent & \warnmark~Medium (preset switch) & \xmark~No (no persistent) & \warnmark~Hybrid inference; cloud auth for advanced \\
      WeChatIME & Hunyuan & Cloud-first & Knowledge retrieval; error correction & \xmark~Weak (general-purpose) & \xmark~No (short-term) & \warnmark~Cloud-dependent; interaction transmission \\
      DoubaoIME & Doubao & Cloud-first & Speech enhancement; association; prediction & \xmark~Weak (no habit learning) & \xmark~No (no persistent) & \warnmark~Cloud + mic access \\
      \system{} (Ours) & IME-specific LLM & On-device & Memory-based completion; persona presets; self-evolving capability & \cmark~Strong (evolving + adaptive) & \cmark~Yes (L1/L2/L3) & \cmark~On-device; auditable \\
      \bottomrule
    \end{tabular}%
    }
    \caption{Comparison of representative LLM-enhanced Chinese IMEs across functionality, personalization, memory, and privacy. We briefly list the key features of different IMEs.}
    \label{tab:comparison}
  \end{table*}

  The first limitation is that generative LLMs are often loosely coupled with the traditional keystroke-to-candidate pipeline, where models are invoked only for long-form rewriting or full-sentence suggestions. This restricts fine-grained modeling of short-phrase completion and interactive edits. Second, personalization mechanisms are shallow, typically confined to short context windows or static personas, without the ability to accumulate long-term user knowledge or continuously adapt from feedback. Third, cloud-dependent inference introduces unpredictable latency and privacy risks for keystroke-level data \cite{hard2018federated, zhao2024privacy}, undermining responsiveness in high-frequency typing scenarios.

  To close this gap, we present \system{}, to the best of our knowledge the first memory-augmented, on-device LLM-driven IME. 
  We tightly integrate the LLM into the IME pipeline, enabling it to play a persistent and central role throughout the typing process, as illustrated in Figure~\ref{fig:teaser}. 
  Through fine-tuning on high-quality synthesized personalization data, we equip the base LLM with initial persona-aware generation capabilities, enabling flexible identity switching.
  To support continual on-device evolution, we design a three-level hierarchical memory architecture together with a GRPO-based memory trigger, enabling the model to autonomously incorporate user-specific signals over time.
  All functions operate efficiently on-device, supported by systemic optimization.Extensive evaluations confirm that our architecture ensures millisecond-level responsiveness and precise, memory-augmented personal writing.

  The contributions of this work are 3-fold:
  \begin{itemize}
      \item We present a synthetic personalization data pipeline and fine-tuning a lightweight base LLM into a persona-aware IME backbone, enabling identity controllable and human-like prediction within the typing loop.
      \item We design a 3-level hierarchical memory architecture, coupled with a GRPO-based trigger, which achieves efficient short-term knowledge recall while supporting long-term user modeling and continual on-device evolution.
      \item We perform systemic hardware-level optimizations tailored to mobile IME workloads, ensuring low-latency and efficient running of LLM-driven IME under real constraints.
  \end{itemize}
  \section{\system{} Design}
  \label{sec:system_design}
  \system{} is designed for extremely resource-constrained mobile environments.
  Our architecture follows four core principles:

  \begin{itemize}[label=\textbullet, leftmargin=1.5em, itemsep=0.2em, topsep=0.3em, parsep=0pt]
    \item \textbf{Seamless integration:} enhance rather than disrupt established keystroke habits.
    \item \textbf{Expressive takeover:} personalized generation with memory-grounded factual augmentation.
    \item \textbf{User-experience first:} co-optimize CPU runtime for millisecond-level responsiveness.
    \item \textbf{Privacy preservation:} keep context and memory on-device rather than online.
  \end{itemize}

  \subsection{Real time interactive UI Design}
  \label{sec:ui_design}
  
  The frontend design of \system{} injects LLM-generated completions into the familiar candidate-based workflow. Figure~\ref{fig:teaser} shows that we use cursor-adjacent \emph{GhostText} with a dual-layer candidate interface.
  GhostText displays the best short-phrase candidate in real time.
  A secondary panel displays alternatives.
  Users accept results by tapping the candidate bubble.
  This preserves existing interaction habits and minimizes behavioral friction.

  \subsection{Stylized LLM Post-Training}
  \label{sec:post_training}

  Generic LLMs tend to produce neutral and impersonal completions. Although such responses are correct, they often lack stylistic consistency and identity-specific expression, resulting in mechanical and context-insensitive candidates. In human-centric communication, however, users type text that reflects personal traits and situational tone. Therefore, stylized post-training is necessary to transform a generic LLM into a persona-aware generator that can produce basic personalized continuations within the typing loop.
 \begin{figure}[!t]
   \centering
   \includegraphics[width=0.98\columnwidth]{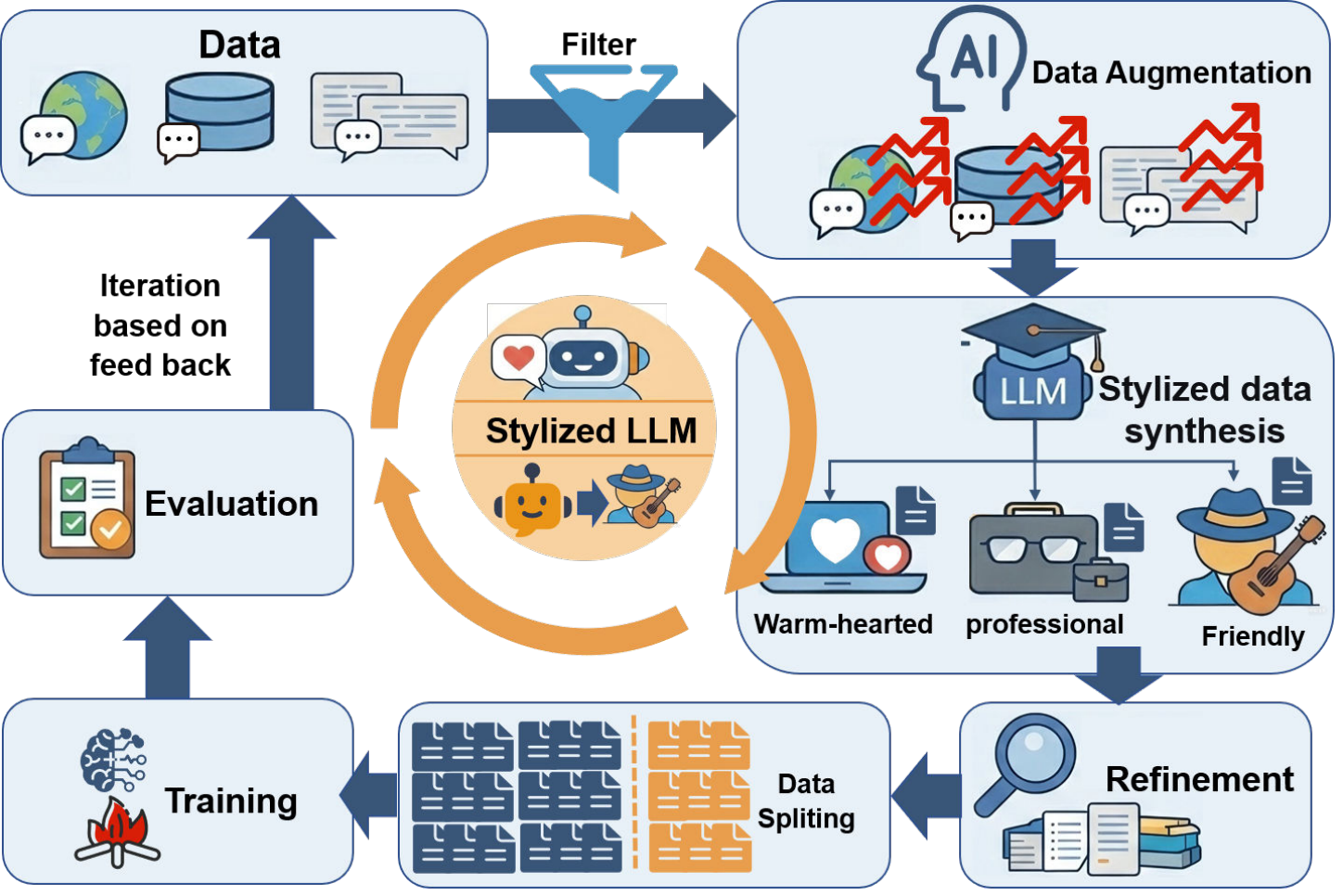}
   \caption{Stylized post-training pipeline. We construct persona-specific conversational data through curation, augmentation, filtering, and iterative refinement.}
   \label{fig:peft}
 \end{figure}

  To this end, we perform stylized post-training over three predefined persona styles, each representing a distinct communicative identity. As shown in Figure~\ref{fig:peft}, we first collect large-scale Internet conversational corpora and augment them with high-quality LLM-synthesized dialogues to enrich scenario diversity. An LLM-assisted filtering and style classification pipeline is then applied to remove low-quality samples and ensure stylistic purity. The curated data are then split into training and test sets. We construct structured training contexts by concatenating personalized system prompt with dialogue history, thereby explicitly generating based on persona. The model is fine-tuned on this stylized corpus and evaluated to assess style fidelity and fluency. Based on evaluation feedback, we iteratively refine data curation and form a closed-loop pipeline that progressively strengthens personality and quality.

  \subsection{Hierarchical Memory Mechanism}
  \label{sec:memory_engine}
  The memory mechanism is the core of \system{}, enabling LLM-based completion to incorporate historical context and better reflect the users's own writing style. Inspired by computer memory hierarchy \cite{packer2024memgpt}, \system{} adopts a decoupled L1/L2/L3 memory architecture separating foreground completion from background curation. They are defined and used as follows:
  \begin{itemize}[label=\textbullet, leftmargin=1.5em, itemsep=0.2em, topsep=0.3em, parsep=0pt]
      \item \textbf{L1 High-speed cache layer:} Stores style-specific and per-memory KV blobs. It directly injects precomputed KV states into the foreground path and reduces prefill overhead.
      \item \textbf{L2 Plaintext fact layer:} Serves as an auditable source of user facts. It appends structured memory records in plaintext and maintains a lightweight HNSW vector index \cite{malkov2018efficient} for semantic recall.
      \item \textbf{L3 Parametric weight:} Accumulates interaction trajectories and model decision logs. It automatically produces datasets for subsequent on-device or online fine-tuning.
  \end{itemize}

  Since performing memory retrieval at every prediction step incurs substantial computational overhead and is often unnecessary, we introduce a GRPO-based memory trigger to selectively activate retrieval when appropriate. In this trigger, we predefine four task classes:
  \textbf{(A) Direct completion} supports foreground typing.
  \textbf{(B) Memory retrieval} emits queries when facts are needed.
  \textbf{(C1) Memory extraction} converts interaction traces into structured records and send them to memory.
  \textbf{(C2) Invalid-information refusal} rejects noisy or sensitive fragments.
  After stylized post-training, the retrieval and refusal boundaries remain unclear. We therefore apply GRPO algorithm~\cite{shao2024deepseekmath} to sharpen these boundaries. The reward function jointly considers format compliance, task correctness, and latency constraints. Detailed GRPO procedures are provided in Appendix~\ref{app:grpo}.

  \subsection{Efficiency Optimization}
  \label{sec:inference_engine}

 To meet millisecond-level latency requirements on heterogeneous mobile CPUs, \system{} adopts a deeply optimized on-device inference runtime that eliminates redundant computation and maximizes cache reuse. We manage the KV cache as a compressed prefix tree \cite{Gusfield:97, zheng2023sglang}, enabling structural sharing across overlapping prefixes caused by typing and edits. Instead of re-prefilling from scratch, the runtime locates the longest common prefix and resumes decoding from the matched node, effectively transforming interactive editing into incremental tree traversal. On big.LITTLE architectures, we further apply thread-affinity scheduling \cite{gerganov2023llama}, dispatching attention and GEMM kernels to high-performance cores, thereby reducing tail latency and stabilizing thermal behavior under bursty workloads.

 For memory-augmented generation, we avoid full re-prefill through a lightweight KV-Splice mechanism that injects precomputed memory KV segments into the active decoding path via position-offset mapping. Since RoPE-based models bind KV states to absolute positions \cite{su2024roformer}, we adopt a position-independent caching strategy inspired by PIC \cite{gim2023promptcache, yang2025kvlink}, using phase-shift correction \cite{yang2025kvlink} together with selective tail recomputation \cite{yao2024cacheblend} to preserve causal attention. This enables flexible cached context stitching with minimal recomputation overhead, maintaining fluent continuation while substantially reducing reinjection cost.

\section{\system{} Implementation}
\label{sec:demo_scenarios}
\label{sec:usage_example}

\begin{figure}[t]
  \centering
  \includegraphics[width=\columnwidth]{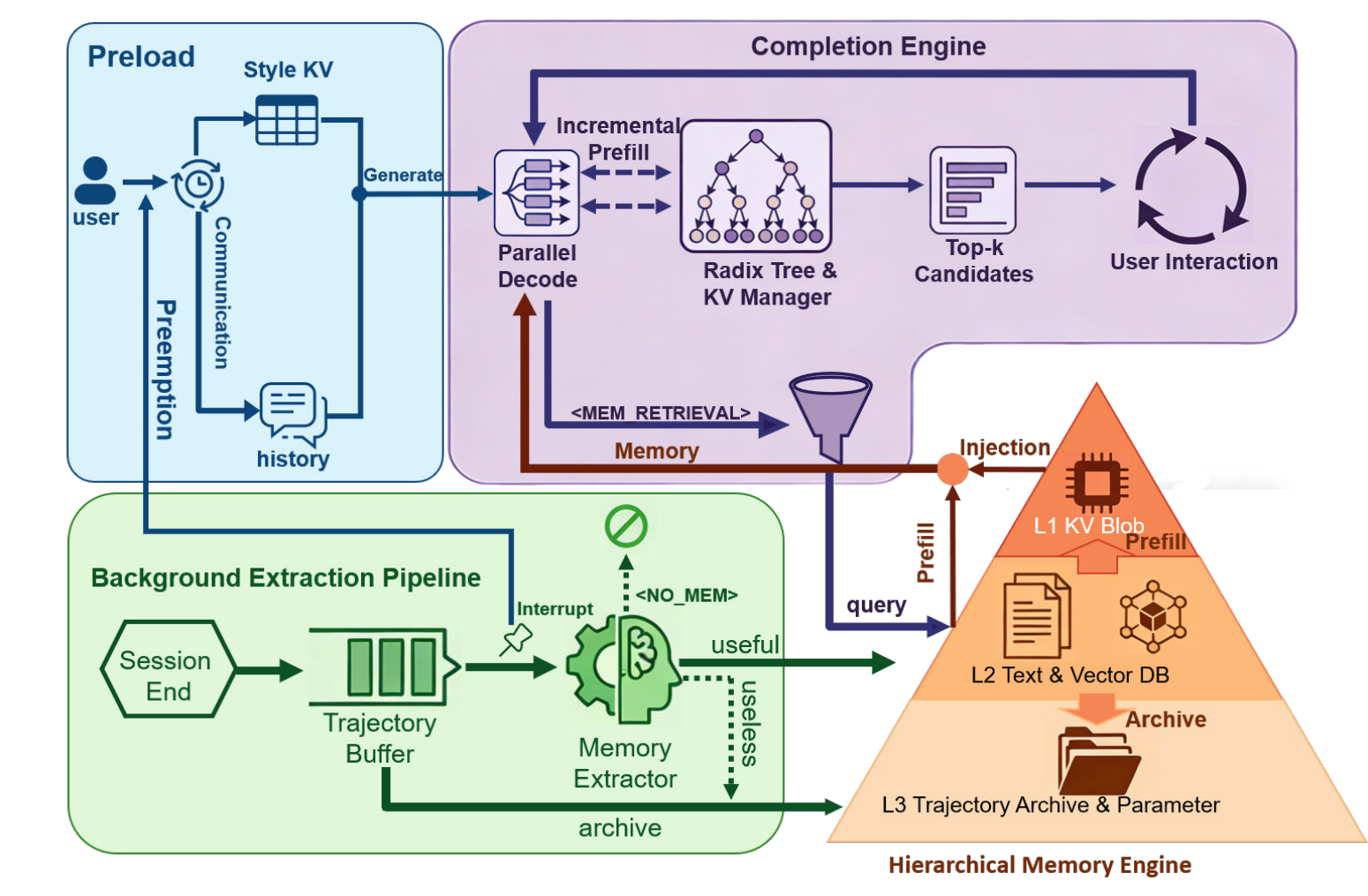}
  \caption{Interaction pipeline between \system{} and HuoziIME-Chat in a daily conversation scenario.}
  \label{fig:usage-flow}
\end{figure}

\Cref{fig:usage-flow} illustrates the end-to-end interaction between \system{} and its companion application \textit{HuoziIME-Chat}. 
We describe a representative daily conversation scenario to demonstrate how memory-augmented short-phrase completion operates under real mobile constraints. 
The implementation tightly couples online inference with asynchronous memory evolution, forming a self-updating personalization loop on device.

\subsection{Online Interaction Loop}

\noindent\textbf{Cross-app context synchronization.}
Due to Android sandbox isolation, real-time context sharing across applications requires explicit coordination.
We implement lightweight cross-process communication via the Model Context Protocol (MCP) \cite{mcp2024}. 
When a chat session changes, authorized host applications issue a \texttt{SYNC} request, enabling dynamic context updates.
If MCP is unavailable, \system{} gracefully degrades to input-only inference without accessing external application context.

Upon receiving a \texttt{SYNC} signal, the engine selects the appropriate L1 style cache and formats dialogue history using the post-training template.
Incremental prefill with RadixTree-based prefix reuse (§2.4) ensures that only modified suffix segments incur computation, minimizing redundant processing.

\noindent\textbf{Memory-augmented candidate generation.}
During typing, the LLM performs multi-threaded decoding to generate short-phrase candidates.
If external factual grounding is required, the model emits the control token \texttt{<MEM\_RETRIEVAL>}, triggering vector search over the local HNSW index.
Retrieved plaintext memories and precomputed KV segments are fused into the active decoding state via KV injection or incremental prefill (§2.4), after which multiple candidate completions are sampled for display.

\noindent\textbf{Incremental interactive rendering.}
The top prediction is rendered as cursor-adjacent GhostText, while alternative candidates appear in the suggestion panel.
As the user continues typing or selecting candidates, the runtime reuses prefix states and performs incremental decoding only on newly appended fragments, amortizing the perceived Time-to-First-Candidate latency to near-zero during interaction.
After message submission, the updated dialogue context is synchronized and the interaction trace is cached for background processing.

\subsection{Asynchronous Memory Update}

\noindent\textbf{Memory extraction and structuring.}
When the conversation ends or the user switches applications, the system transitions to a background curation phase.
Cached interaction traces are converted into structured records (L3), and the GRPO-trained policy determines whether factual extraction or refusal is appropriate.
Candidate facts are rewritten into concise declarative form for persistent storage.

\noindent\textbf{Tiered persistence with strict preemption.}
Validated memories are inserted into the L2 plaintext store, indexed via HNSW, and optionally compiled into reusable L1 KV blobs.
To preserve real-time responsiveness, strict foreground preemption is enforced: any incoming \texttt{SYNC} request immediately interrupts background execution and restores the online inference path.

Together, the online and asynchronous pipelines enable continual on-device personalization under strict mobile constraints.
\section{Experiments}
\label{sec:evaluation}

We evaluate \system{} on an Android test device equipped with a MediaTek Dimensity 9000 SoC and 12\,GB RAM.
Our evaluation focuses on memory retrieval effectiveness and on-device inference performance to validate its usability as a real-time mobile application.

\subsection{Memory Pipeline Effectiveness}

We evaluate the memory pipeline across its complete lifecycle, i.e., Triggering, Processing, Retrieval, and Grounded Generation. Specifically, the \textit{Triggering} stage uses a targeted input prefix to prompt the model to initiate a retrieval request. \textit{Processing} handles the frontend user input, either extracting and storing relevant facts into memory or actively disregarding noisy, irrelevant context. The \textit{Retrieval} stage queries the memory database to fetch the most pertinent information. Finally, \textit{Grounded Generation} assesses the model's ability to seamlessly incorporate the retrieved memory into an output that is both factually accurate and stylistically coherent. The test cases are drawn from our pre-synthesized dataset.

\begin{table}[t]
    \centering
    \small
    \begin{tabular}{lcc}
        \toprule
        \textbf{Pipeline Stage} & \textbf{Success / Total} & \textbf{Rate (\%)} \\
        \midrule
        Memory Trigger & 342 / 343 & 99.7 \\
        Processing (Normal) & 163 / 169 & 96.4 \\
        Processing (Refusal) & 87 / 122 & 71.3 \\
        \midrule
        Retrieval@4 & 179 / 200 & 89.5 \\
        \midrule
        Grounded Generation & 156 / 179 & 87.2 \\
        \bottomrule
    \end{tabular}
    \caption{Performance of the memory triggering and retrieval effectiveness.}
    \label{tab:memory_metrics}
\end{table}

As presented in Table~\ref{tab:memory_metrics}, the evaluation demonstrates that, beyond stylized text completion, our model successfully masters memory operations. Delving into the empirical metrics, normal processing successfully extracts factual content into valid, constraint-compliant formats in 96.4\% of cases. For the stricter refusal task, the model accurately outputs a \texttt{<NO\_MEM>} token to discard meaningless context and prevent memory database pollution with a 71.3\% success rate. During retrieval, we query a local vector database of 200 memories using the quantized \texttt{bge-small-zh-v1.5} model, evaluating exact alignments against gold-truth IDs. By leveraging the Retrieval@4 metric, which directly determines the default four-candidate UI layout of \system{}, we achieve a strong hit rate of 89.5\% (179/200). This highlights a core system design advantage: the multi-candidate nature of the IME naturally compensates for the inherent limitations of lightweight, on-device embedding models. Building upon these successfully retrieved cases, manual human inspection confirms an 87.2\% (156/179) success rate for grounded generation, demonstrating that \system{} reliably synthesizes retrieved knowledge into stylistically coherent keystroke predictions suitable for real-time user interaction.

\subsection{On-Device Inference Performance}
\begin{figure*}[t]
    \centering
    \begin{subfigure}[t]{0.43\textwidth}
        \centering
        \includegraphics[width=\linewidth]{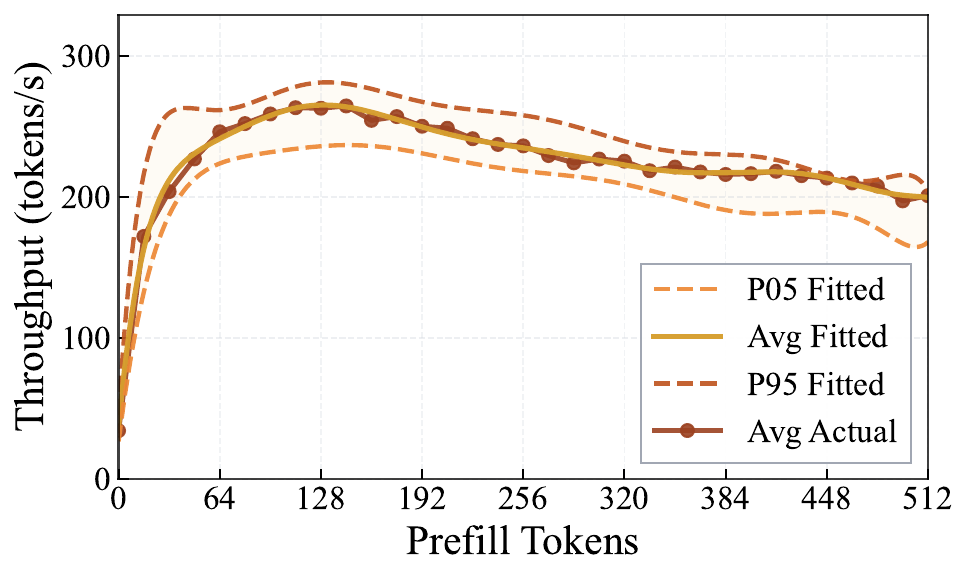}
        \caption{Prefill Throughput}
        \label{fig:eval_prefill}
    \end{subfigure}
    \hspace{10mm}
    \begin{subfigure}[t]{0.43\textwidth}
        \centering
        \includegraphics[width=\linewidth]{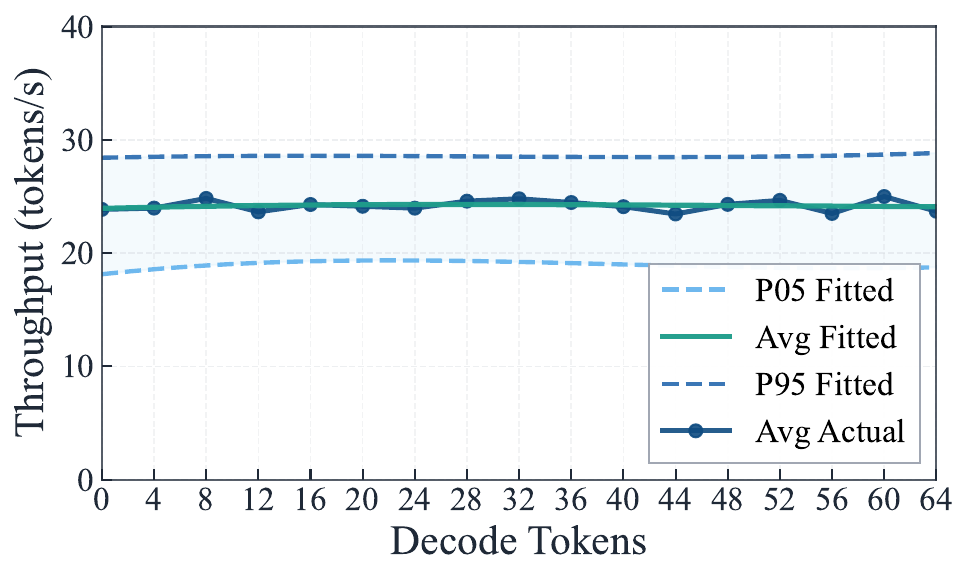}
        \caption{Decode Throughput}
        \label{fig:eval_decode}
    \end{subfigure}

    \vspace{0.2em}
    \begin{subfigure}[t]{0.43\textwidth}
        \centering
        \includegraphics[width=\linewidth]{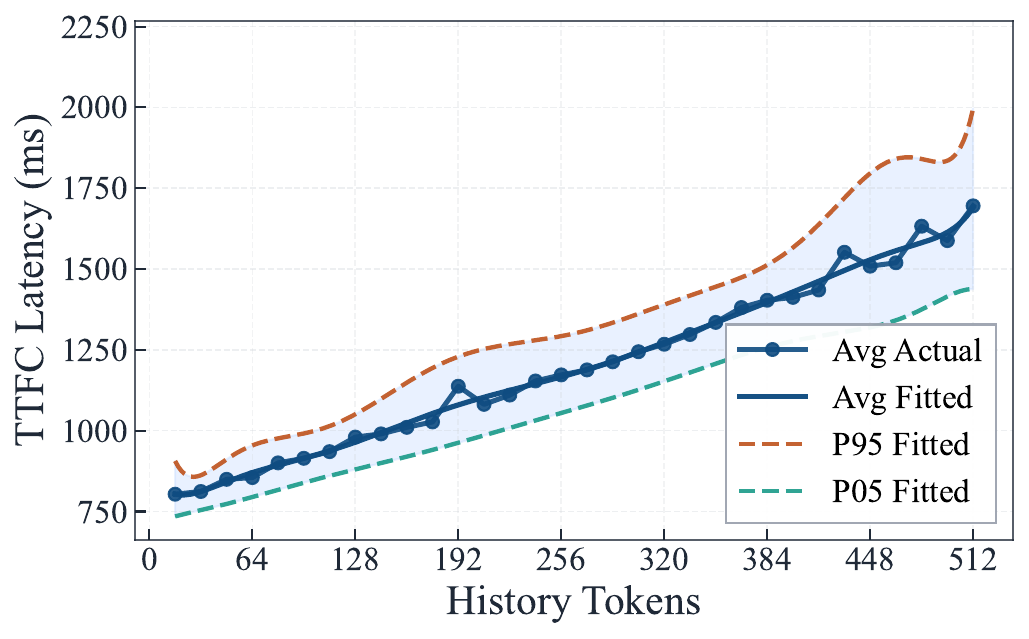}
        \caption{TTFC Latency}
        \label{fig:eval_ttfc}
    \end{subfigure}
    \hspace{12mm}
    \begin{subfigure}[t]{0.43\textwidth}
        \centering
        \includegraphics[width=\linewidth]{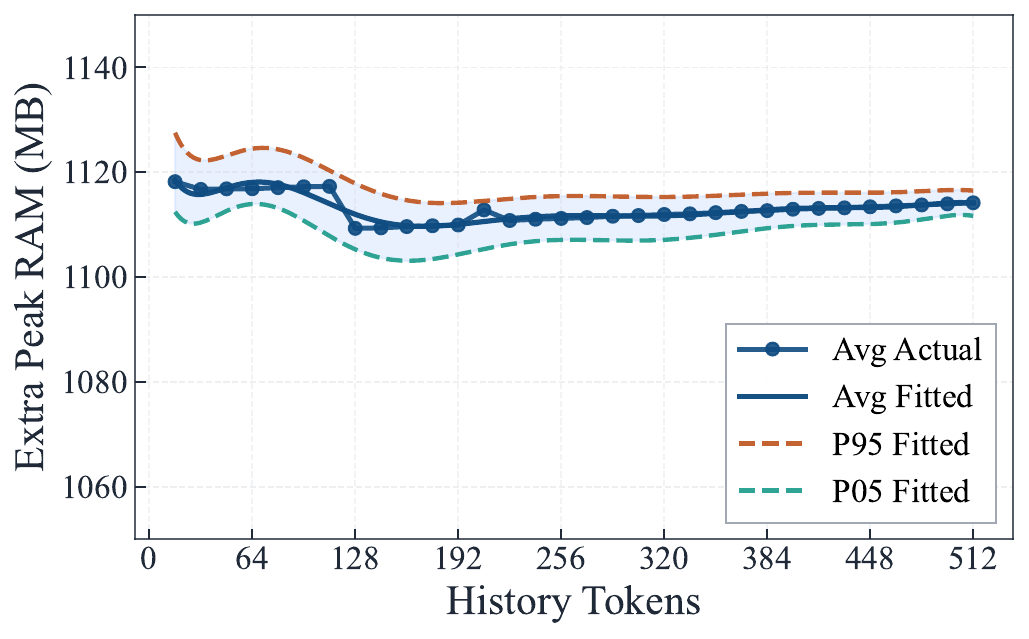}
        \caption{Extra Peak RAM}
        \label{fig:eval_ram}
    \end{subfigure}
    \caption{On-device inference performance across context lengths up to 512 tokens: (a) prefill throughput, (b) decode throughput, (c) TTFC latency, and (d) extra peak RAM.}
    \label{fig:eval_overall}
\end{figure*}

To ensure a seamless typing experience, we profile the core inference metrics across varying context lengths up to 512 tokens, as shown in Figure~\ref{fig:eval_overall}.

\paragraph{Throughput \& Latency.}

Benefiting from our thread-affinity scheduling (\S\ref{sec:inference_engine}), \system{} achieves a peak prefill throughput of over 260 token/s.
During the autoregressive phase, the decode throughput remains remarkably stable at 24--25 tokens/s, which comfortably outpaces standard human reading and typing speeds to ensure fluid \emph{GhostText} rendering.
Although the cold-start Time to First Candidate (TTFC) scales linearly from 800\,ms to 1700\,ms, continuous interactive typing largely avoids this overhead. By leveraging RadixTree-based KV reuse and KV-Splice stitching, \system{} bypass re-prefilling and supports efficient incremental decoding. This amortizes the perceived TTFC to near-zero, preserving the user's cognitive flow.

\paragraph{Memory Overhead.}

To avoid out-of-memory, we quantize the model from 1.34\,GB to 485\,MB (Q4\_0) and bound the L1 KV cache to 24\,MB. Figure~\ref{fig:eval_ram} shows dynamic peak RAM safely stabilizes at $\sim$1.12\,GB, ensuring seamless background co-existence with heavy foreground apps.

\subsection{User Feedback}

We invite four participants with NLP backgrounds to engage in free multi-turn conversations using \system{}. Participants report that cross-session recall and context-aware styling noticeably reduce typing friction and improve the overall writing experience, consistent with gains in Keystroke Savings Rate (KSR).

\section{Related Work}
  IMEs are initially dominated by transliteration-based methods, with generative paradigms emerging as the field advanced.
  Based on N-gram models, early IMEs strictly map keystrokes to text sequences. Afterwards, RNN-based IMEs~\cite{chen2015neural,huang2018moon,hard2018federated} endow them with certain generative capabilities. However, they inherently lack deep user intent modeling, cross-application context, and true stylistic personalization beyond static lexicons.
  
  Driven by LLMs, the objective shifts from deterministic mapping to next-phrase generation \cite{zhao2023survey,bubeck2023sparks,qwen2023}. These systems leverage agentic capabilities, such as memory and tool use, to continuously enhance personalization and generation quality \cite{packer2024memgpt,mcp2024}. To our knowledge, \system{} is the first generative Chinese IME in our target setting. It addresses the fundamental quality-efficiency-privacy trilemma by deploying a unified post-trained model for both generation and decision-making \cite{zhao2024privacy}. Through algorithmic and hardware co-optimization, alongside strictly localized data processing, \system{} provides an exploratory yet robust solution \cite{hard2018federated,qwen2023}.

\section{Conclusion}
We presented \system{}, the first fully on-device, memory-augmented generative Chinese IME. By integrating a persona-adapted LLM, a GRPO-enhanced hierarchical memory, and MCP-based cross-app communication, it transforms static typing into a proactive, personalized experience. Our deeply optimized CPU runtime ensures near-zero latency and a strict small memory footprint. Evaluations confirm \system{} delivers stable on-device deployment and user feedback.

For future work, we will focus on customized pre-training to enforce stronger agentic awareness and structural formatting. We will also incorporate more advanced edge-side inference optimizations to push the reliability and quality ceilings under extreme hardware constraints.
\section*{Limitations}

While \system{} demonstrates the feasibility of an on-device generative IME, several limitations remain. The reasoning capability is inherently bounded by the lightweight model size, leading to occasional agentic instabilities such as over-retrieval and retrieval-token drift under autoregressive sampling. In addition, efficiently and securely acquiring real-time external application context remains challenging due to sandboxing constraints if mobile operating system.

  \section*{Ethics Statement}
\system{} adopts an on-device deployment paradigm to minimize cloud-side privacy risks and provides transparent local workflows for data collection and deletion. Nevertheless, locally stored user traces may remain vulnerable if a device is compromised, and robust system-level access protection is essential. The system is built upon the Qwen3-0.6B series \cite{qwen2023} with secondary development of \texttt{llama.cpp} and YuyanIME under GPL-3.0. Although the training corpus undergoes automated cleaning and debiasing, residual bias may persist, and user supervision is recommended in practice. Preliminary user feedback is collected from volunteer teammates, with only anonymized identifiers recorded and no personally identifiable information retained.

\bibliography{custom}

\appendix
\section{Appendix}
 \subsection{Reward in the GRPO Algorithm}
  \label{app:grpo}

  The GRPO algorithm optimizes the trainable policy $\pi_\theta$ by sampling a group of $G$ outputs for each input prompt $x$ from the behavior policy and estimating their relative advantages via in-group reward normalization. The policy is then updated by minimizing a PPO-style clipped surrogate objective \cite{schulman2017proximal}, regularized by a KL-divergence penalty against the frozen reference policy $\pi_{\text{ref}}$. The core driver of these optimization updates is the rule-based reward score $r(x,y)$, which dictates the advantage estimations for each generation.

  
  The total reward is the sum of a thinking-format term and a task-content term:
  \begin{equation}
  r(x,y)=r_{\text{think}}(y)+r_{\text{task}}(x,y).
  \end{equation}
  where $r_{\text{think}}$ is the thinking-format reward and $r_{\text{task}}$ is the task-content reward. Let $L_{\text{think}}$ be the thought-content length and let $I_{\text{tags}}$ denote tag integrity, we have the following definition:
  
  {\small
  \begin{equation}
  r_{\text{think}}(y)=
  \begin{cases}
    +0.2 & \text{if valid think and } 0<L_{\text{think}}\le 300,\\
    -0.2 & \text{if } L_{\text{think}}>300 \text{ or no body text},\\
    -0.5 & \text{if } I_{\text{tags}}=\text{Broken},\\
    0.0  & \text{otherwise.}
  \end{cases}
  \end{equation}}
  
  
  Task scores are computed by prompt type $T$, i.e., $A, B, C1$ or $C2$. For \textbf{normal continuation} ($T=A$), the reward is defined as follows:
  
  {\small
  \begin{equation}
  r_{\text{task}}=
  \begin{cases}
    -1.5 & \text{if mem-retrieval tag is emitted},\\
    1.5-P_{\text{qual}} & \text{otherwise.}
  \end{cases}
  \end{equation}}

  Here $P_{\text{qual}}$ denotes the quality penalty. For \textbf{memory retrieval} ($T=B$), the reward is defined as follows:

  {\small
  \begin{equation}
  \begin{aligned}
  r_{\text{task}}=
  S_{\text{fmt}}+
  0.5\cdot\mathbb{I}(q\neq\emptyset)\\
  &-
  1.0\cdot\mathbb{I}(\text{text}\notin\text{tags}),
  \end{aligned}
  \end{equation}}
  
  Here $S_{\text{fmt}}$ is the format score and
  
  {\small
  \begin{equation}
  S_{\text{fmt}}\in\{2.5\ \text{(perfect)},\ 0.3\ \text{(partial)},\ -1.0\ \text{(wrong)}\}.
  \end{equation}}

  For \textbf{memory extraction} ($T=C1$), the reward is defined as follows:

  {\small
  \begin{equation}
  r_{\text{task}}=
  \begin{cases}
    -1.5 & \text{if pure NO-MEM is emitted},\\
    -1.0 & \text{if JSON is invalid},\\
    1.5+0.2\cdot N_{\text{fields}} & \text{if JSON is valid.}
  \end{cases}
  \end{equation}}
  
  The last term adds a richness bonus proportional to $N_{\text{fields}}$. For \textbf{extraction refusal} ($T=C2$), the reward is defined as follows:

  {\small
  \begin{equation}
  r_{\text{task}}=
  \begin{cases}
    1.5 & \text{if pure NO-MEM is emitted},\\
    -1.0 & \text{if } y \text{ generates JSON},\\
    0.0 & \text{if noisy NO-MEM appears with extra text.}
  \end{cases}
  \end{equation}}

  \subsection{Details of the KV-Splice Algorithm}

In Algorithm~\ref{alg:kv-splice}, we present the detailed procedure of the KV-Splice mechanism introduced in Section~\ref{sec:inference_engine}. To avoid the high computational overhead of full context re-prefilling when injecting retrieved memory facts ($M$) into the active decoding path, KV-Splice operates by identifying the common prefix ($P$) and suffix ($S$) between the stable base sequence and the memory-augmented sequence. Since RoPE-based models strictly bind KV states to absolute positions, directly inserting new tokens would invalidate the positional encodings of all subsequent tokens. To resolve this, the algorithm calculates a position-offset mapping ($\Delta$) and applies phase-shift correction to the retained suffix KV states. It then selectively recomputes only the newly injected memory tokens and the final tail token to seamlessly preserve causal attention. Furthermore, to ensure inference robustness under extreme mobile constraints, the algorithm incorporates a fallback mechanism that safely reverts to a standard baseline prefill if the resulting KV states are non-consecutive or fail to yield valid candidates.
  
  \begin{algorithm}[H]
  \caption{KV-Splice for Phrase Candidate Generation}
  \label{alg:kv-splice}
  \begin{algorithmic}[1]
    \Require Stable base sequence $\texttt{seq}_0$, prompt segments $(P,M,S)$, candidate count $K$
    \Ensure Candidate set $\mathcal{Y}$
    \State $\mathbf{b} \gets \textsc{Tok}(P \Vert S)$, $\mathbf{f} \gets \textsc{Tok}(P \Vert M \Vert S)$
    \State $(\ell_p,\ell_s) \gets \textsc{CommonPrefixSuffix}(\mathbf{b},\mathbf{f})$
    \State $p_{\mathrm{ins}} \gets \ell_p$
    \State $d \gets |\mathbf{b}| - \ell_p - \ell_s$ \Comment{deleted span length in base}
    \State $n \gets |\mathbf{f}| - \ell_p - \ell_s$ \Comment{inserted span length from memory}
    \State $\Delta \gets n-d$
    \State $s_w \gets n_{\mathrm{seq,max}}-2,\;\; s_t \gets n_{\mathrm{seq,max}}-1$
    \State $\textsc{SeqCp}(\texttt{mem}, \texttt{seq}_0, s_w, 0, -1)$
    \State $\textsc{SeqRm}(\texttt{mem}, s_w, p_{\mathrm{ins}}, p_{\mathrm{ins}}+d)$
    \State $\textsc{SeqAdd}(\texttt{mem}, s_w, p_{\mathrm{ins}}+d, -1, \Delta)$
    \State $\textsc{SeqCp}(\texttt{mem}, \texttt{seq}_0, s_t, 0, -1)$
    \State $\textsc{SeqRm}(\texttt{mem}, s_t, p_{\mathrm{ins}}, -1)$ \Comment{keep prefix only}
    \For{$j=0$ to $n-1$}
      \State $x \gets \mathbf{f}[p_{\mathrm{ins}}+j]$
      \State $\textsc{Decode1}(x,\mathrm{pos}=p_{\mathrm{ins}}+j,\mathrm{seq}=s_t,\mathrm{logits}=0)$
    \EndFor
    \State $\textsc{SeqCpOverlay}(\texttt{mem}, s_t, s_w, p_{\mathrm{ins}}, p_{\mathrm{ins}}+n)$
    \State $p_{\mathrm{last}} \gets |\mathbf{f}|-1,\;\; x_{\mathrm{last}} \gets \mathbf{f}[p_{\mathrm{last}}]$
    \State $\textsc{SeqRm}(\texttt{mem}, s_w, p_{\mathrm{last}}, p_{\mathrm{last}}+1)$
    \State $\textsc{Decode1}(x_{\mathrm{last}},\mathrm{pos}=p_{\mathrm{last}},\mathrm{seq}=s_w,\mathrm{logits}=1)$
    \State $\mathcal{Y} \gets \textsc{SampleCandidates}(s_w, K)$
    \If{$\mathcal{Y}=\varnothing$ \textbf{or} $\neg\textsc{PosConsecutive}(s_w)$}
      \State $\mathcal{Y} \gets \textsc{Baseline}(P,M,S,K)$
    \EndIf
    \State \Return $\mathcal{Y}$
  \end{algorithmic}
  \end{algorithm}

\end{document}